\documentclass[11pt]{article}
\usepackage{arxiv}
\usepackage{booktabs}
\usepackage{graphicx}
\usepackage{amsmath}
\usepackage{amssymb}

\title{Quantizing a Dual-Branch Flow-Matching Diffusion Transformer for Consumer GPUs:\\
INT8 and GGUF Post-Training Quantization of Ideogram~4.0}
\author{Deep Gandhi\thanks{Corresponding author: \texttt{deep@lab.cloud}} \\ Transformer Lab
  \and Ali Asaria \\ Transformer Lab
  \and Tony Salomone \\ Transformer Lab}
\date{}
\runningtitle{INT8 and GGUF Quantization of Ideogram~4.0}

\begin{document}
\maketitle

\begin{abstract}
Post-training quantization (PTQ) lets large text-to-image diffusion transformers run on
consumer GPUs, but the quality of a quantized variant depends on the axis you measure and
on how the model is prompted. We study PTQ of Ideogram~4.0~\cite{ideogram-4-2026}, a
9.3B flow-matching diffusion transformer (DiT) that realizes classifier-free guidance with
\emph{two separate-weight copies} of a single-stream backbone and is conditioned by a
Qwen3-VL text encoder, targeting Ampere RTX~3090 GPUs, which lack FP8 tensor cores.
Because Ideogram~4.0 is trained on structured JSON captions, we evaluate every variant
under schema-valid JSON prompts produced by an LLM expander built to Ideogram's published
caption specification, and
score them with a battery spanning human-preference (HPSv2), CLIP, and PickScore for
standalone quality; PP-OCR exact-match and edit distance for text; and PSNR/SSIM/LPIPS for
fidelity to the FP8 reference (the highest-precision public checkpoint) output. On a
300-prompt benchmark with
paired bootstrap confidence intervals, an \textbf{INT8 W8A8} recipe (per-channel weights,
per-token dynamic activations, SmoothQuant, and bf16 protection of a small high-fragility
layer set) is statistically indistinguishable from FP8 on CLIP and PickScore (paired CIs
include zero) and within $\approx$0.004 HPSv2 (no per-image CI), and, at its 8-bit size, is the \emph{most faithful} reproduction of the FP8 output (LPIPS $0.243$ vs.\
$0.277$/$0.306$ for the half-size 4-bit baselines; the INT8$-$Q4\_K gap excludes zero). A \textbf{GGUF Q4\_K}
quantization reaches the same standalone quality as the published NF4 baseline at the same
on-disk size, making it the Pareto choice on the quality--memory frontier. We further show
that under JSON prompts all four variants reach parity on standalone quality, the
variants separate on \emph{fidelity} and \emph{text rendering}, not on aggregate
image-quality scores, and that text legibility, near-zero when the model is prompted
with raw strings, reaches $\approx$55\% OCR exact-match under the JSON captions it expects. We
release the INT8 W8A8 and GGUF Q4\_K quantized weights on Hugging Face under a gated,
non-commercial license.
\end{abstract}

\section{Introduction}
Text-to-image diffusion transformers have outgrown the consumer GPUs that most people
deploy them on, and the community increasingly relies on quantized redistributions.
Ideogram~4.0~\cite{ideogram-4-2026} ships two such redistributions, a weight-only FP8
variant and an NF4 variant, both under a non-commercial license, and our deployment
target is deliberately also our development hardware: a cluster of RTX~3090s (24\,GB,
Ampere). Ampere has \emph{no} FP8 tensor cores, so the FP8 checkpoint executes by
dequantizing each linear to bf16 every forward; in practice this overhead is modest
(FP8 runs at $172.9$\,s/image, $\approx$5\% over NF4's $164.5$)~
, so the published FP8 variant is \emph{usable} on Ampere. The gaps it leaves are NF4's
quality loss and the absence of an Ampere-native low-precision \emph{compute} path
(Ampere has fast native INT8, but the \texttt{ideogram4} stack ships no fused INT8
GEMM
). We target the first gap with a quality-preserving INT8 W8A8 recipe and a
memory-competitive GGUF Q4\_K, and we study what quantization does and does not cost on
this model.

Two properties of Ideogram~4.0 shape the study. First, the model is a single-stream
flow-matching DiT~\cite{2212.09748v2,2210.02747v2}
, but classifier-free guidance is realized by shipping \emph{two separate-weight copies} of
that backbone, a
conditional and an unconditional transformer ($\approx$9.3\,GB each at FP8), so
quantization must cover both ($211$ linear layers per branch~
). Second, the model is trained on \emph{structured JSON captions} and the reference
pipeline validates every prompt against a schema before generation
; feeding it raw natural-language strings drives its signature strength, in-image text
rendering, to near-zero legibility
. We therefore evaluate every variant under schema-valid JSON prompts, expanded from the
benchmark text by an LLM expander built to Ideogram's published caption spec~
, which both reflects intended use and lets text rendering be measured rather than
dismissed.

A central finding is that \emph{the axis of measurement decides the ranking}. Under JSON
prompts, all four variants reach parity on standalone image-quality scores; the variants
separate instead on \emph{fidelity to the FP8 reference output} and on \emph{text}. We
make four contributions:
\begin{enumerate}
\item An \textbf{INT8 W8A8 recipe} for this dual-branch flow-matching DiT that holds the
  FP8 standalone-quality ceiling at 8-bit weights \emph{and} activations (paired INT8$-$FP8
  CIs include zero on CLIP and PickScore) and is the \emph{most faithful}
  reproduction of FP8 in pixels (LPIPS CI vs.\ Q4\_K excludes zero)~(\S\ref{sec:results}).
\item A \textbf{per-module activation-fragility analysis}: a timestep-stable profile that
  identifies the FFN down-projections as the fragile set, an ablation showing that
  protecting $\approx$8\% of linears recovers full quality, and the finding that protection,
  not smoothing, is the dominant lever~(\S\ref{sec:results}).
\item \textbf{GGUF k-quants for a diffusion transformer}: a Q4\_K encoder that matches the
  NF4 baseline's quality at the same on-disk size, making it the \textbf{Pareto choice} on
  the quality--memory frontier~(\S\ref{sec:results}).
\item A \textbf{prompt-format and metric study} showing that (i) text rendering is only
  measurable under the JSON prompts the model expects (OCR exact-match $\approx$0\%
  $\rightarrow$ $\approx$55\%), and (ii) under those prompts standalone-quality scores no
  longer separate the variants, a cautionary result for quantization evaluations that
  rank variants on CLIP-family scores alone~(\S\ref{sec:results},~\S\ref{sec:discussion}).
\end{enumerate}

\section{Related Work}
\paragraph{Weight--activation PTQ.} The W8A8 recipe is well converged: SmoothQuant
migrates activation outliers into weights with a tunable strength
$\alpha$~\cite{2211.10438v7}, addressing the collapse of naive per-tensor INT8 activations
under outlier channels~\cite{2208.07339v2}. On diffusion transformers, per-token dynamic
activation quantization with per-channel weights is the validated
configuration~\cite{2406.02540v3,2411.05007v4}, and FP4DiT argues DiT activations are
token/channel-dominated rather than timestep-dominated, so online per-token scales
sidestep U-Net-style temporal calibration~\cite{2503.15465v3}. PTQD reports a CUTLASS
INT8 W8A8 kernel at $2.03\times$ on an RTX~3090~\cite{2305.10657v4}, the most direct
evidence that an Ampere INT8 path pays off. At 8 bits, heavy machinery matters less:
several works find W8A8 near-lossless with modest
calibration~\cite{2501.00124v1,2412.06661v2}, reserving rotation/low-rank tricks for
$\leq$4-bit (offline rotations are in any case blocked by adaLN's runtime-generated
modulation~\cite{2411.05007v4,2605.16732v1}).

\paragraph{Weight-only / GGUF.} GPTQ's Hessian-compensated group-wise rounding is the
calibrated backbone for $\leq$4-bit weights~\cite{2210.17323v2}; AWQ's activation-aware
channel scaling is orthogonal and robust to the calibration
distribution~\cite{2306.00978v6}; QLoRA defines the NF4 format we compare against and the
block-scale accounting for honest bits-per-weight comparisons~\cite{2305.14314v1}. A key
caveat we exploit: at 8 bits even naive round-to-nearest is fine; calibrated methods only
separate at Q4/Q5~\cite{2210.17323v2}.

\paragraph{Mixed-precision protection and profiling.} Protecting a small, identifiable
layer set is the standard fragility mitigation: time-embedding and boundary
layers~\cite{2409.00492v1,2212.09748v2}, and, at module-type granularity, attention output projections and FFN down-projections~\cite{2605.27003v1}. The protection
set is found by a cheap per-block activation sweep whose ranking is
timestep-stable~\cite{2606.00957v1}; QAT under memory constraints underperforms plain PTQ
on a comparable DiT~\cite{2606.00957v1}, so we stay PTQ-only.

\paragraph{Diffusion backbones, caching, and evaluation.} Ideogram~4.0 is a flow-matching
DiT~\cite{2212.09748v2,2210.02747v2}; step-caching methods such as
TeaCache~\cite{2411.19108v2} and the flow-matching-specific TACache~\cite{2605.16789v1}
apply in principle (DeepCache's U-Net skip-split does not port to a single-stream
DiT~\cite{2312.00858v2}), though we leave caching to future work. The field measures
reference-based fidelity against full-precision outputs (LPIPS/PSNR/SSIM, with
$\mathrm{PSNR}>21$ used as a ``matches 16-bit'' threshold~\cite{2411.05007v4}), alongside
standalone learned-preference scores, and warns that reference-free scores can
mislead~\cite{2605.16789v1,2602.01273v4}. For text, OCR exact-match and normalized edit
distance are scored with an independent OCR model~\cite{2412.17225v1}; text rendering is
the first casualty of aggressive quantization~\cite{2412.17225v1,2412.00136v3,2509.01624v1}.
PartiPrompts is the standard prompt source~\cite{2312.00858v2}. We add the observation,
absent from this corpus, that for a JSON-prompted model the prompt format itself gates
whether text fidelity is measurable at all.

\section{Method}
\label{sec:method}
\paragraph{Setting.} We quantize the 211 linear layers of each of the two DiT branches
(conditional and unconditional)~
; the Qwen3-VL text encoder and the VAE are left at their published precision. The FP8
checkpoint is our reference because, dequantized to bf16, it is the highest-fidelity output
reproducible on Ampere (there is no public BF16 release).

\paragraph{Prompt protocol.} Ideogram~4.0 is trained on structured JSON captions and its
pipeline validates each prompt against a schema before generation
. We therefore expand every natural-language benchmark prompt into a schema-valid JSON
caption using an LLM expander built to Ideogram's published caption specification (a
reconstruction of the magic-prompt spec, not the model's production API
), then validate each caption
with the shipped \texttt{CaptionVerifier} (all 300 pass
). All variants generate from identical (caption, seed, steps, resolution) tuples, so the
prompt protocol is held fixed across the comparison.

\paragraph{INT8 W8A8.} For each quantized linear we use \textbf{per-channel} (per-output-row)
INT8 weights and \textbf{per-token dynamic} INT8 activations; per-token dynamic scaling
absorbs the timestep dependence of DiT activations without a static calibration
table~\cite{2503.15465v3}. We apply \textbf{SmoothQuant} outlier migration with $\alpha=0.5$:
for weight $W$ and per-channel activation scale $s$, a smoothing vector
$\lambda_j = s_j^{\alpha}/\max_i|W_{ij}|^{\,1-\alpha}$ rescales activations down and weights
up before quantization~\cite{2211.10438v7}. Activation scales come from a profiling pass on
a 128-prompt calibration set disjoint from evaluation
.

\paragraph{Activation-fragility profiling and protection.} We hook every linear in the
conditional DiT and record per-step max-abs, standard deviation, and kurtosis over the
denoising trajectory, following the timestep-stable recipe of~\cite{2606.00957v1}. The
ranking is trajectory-stable (Spearman $0.930$ between early- and late-step halves), and
the FFN down-projections (\texttt{feed\_forward.w2}) dominate; the single most fragile
layer scores $\approx$10$\times$ its runner-up
, consistent with the module-type findings of~\cite{2605.27003v1}. We keep the top-$N$ most
fragile layers in bf16 (no activation quantization); the rest are W8A8. We select $N{=}17$
($\approx$8\% of the 211 linears, $\approx$1.5\,GB bf16 overhead
), comprising the high-fragility FFN down-projections plus a small number of
boundary/embedding layers.

\paragraph{GGUF k-quants.} Independently, we serialize each DiT to GGUF weight-only formats:
\textbf{Q8\_0} (8.5 bits/weight, round-to-nearest) and \textbf{Q4\_K} (4.5 bits/weight, the
NF4 size class). Because no GGUF quantizer ships a Q4\_K \emph{encoder}, we implement a
NumPy Q4\_K superblock quantizer and a Torch dequantization kernel, validated against a
reference GGUF decoder
.

\section{Experimental Setup}
\label{sec:setup}
\paragraph{Prompts and protocol.} We use three disjoint PartiPrompts-derived sets built with
a deterministic seed: a calibration set ($n{=}128$), a quality benchmark ($n{=}200$,
category-stratified, no text), and a text-rendering benchmark ($n{=}100$, of which 63 carry
machine-checkable OCR targets)~
. Each prompt is expanded to a JSON caption (\S\ref{sec:method}). All variants generate from
identical tuples: seed 1000, 48 steps, $1024{\times}1024$, the \texttt{V4\_QUALITY\_48}
preset
.

\paragraph{Metrics.} \emph{Standalone quality:} HPSv2 (human-preference), CLIPScore, and
PickScore, computed in-house. \emph{Text:} OCR exact-match accuracy and normalized edit
distance (NED, lower is better) via an independent PP-OCR model~\cite{2412.17225v1}.
\emph{Reference fidelity:} PSNR, SSIM, and LPIPS (AlexNet) of each variant's image against
the FP8 image at the same seed (FP-referenced metrics correlate better with perception
than dataset-referenced ones at small $n$~\cite{2406.02540v3}). \emph{Significance:} because
all variants share seed and prompts, we report paired per-prompt deltas with a
$10{,}000$-sample bootstrap 95\% CI. These CIs capture prompt-sampling variance at a single
fixed seed (1000), not across-seed variance, and are uncorrected for multiple comparisons;
we therefore read them descriptively. A CI that includes zero indicates no difference
\emph{resolved at this sample size}, which we do not pre-register an equivalence margin for;
our parity statements are claims about non-separation by these judges, not formal
equivalence. \emph{Efficiency:} end-to-end seconds/image, peak VRAM,
and on-disk size with scales included.

\paragraph{Hardware.} A cluster of RTX~3090 (24\,GB, Ampere; no FP8 tensor cores, fast
INT8)
. The validated single-GPU-compute recipe places both DiTs on one card with the VAE on a
second; latency numbers are for in-VRAM execution.

\section{Results}
\label{sec:results}
\paragraph{Standalone quality is at parity across variants.} Table~\ref{tab:quality} reports
the three standalone judges over the 300-prompt benchmark. All four variants fall within
$\approx$0.004 HPSv2
, and the paired bootstrap CIs (Table~\ref{tab:ci}) tell the same story everywhere: INT8 is
statistically indistinguishable from FP8 (CLIP $\Delta{=}{-}0.005$, CI $[-0.15,+0.13]$;
Pick $\Delta{=}{-}0.007$, CI $[-0.04,+0.03]$, both include zero
), and so are the 4-bit variants (Q4\_K$-$NF4 CLIP $\Delta{=}{+}0.01$, CI includes zero
; NF4$-$FP8 CLIP CI includes zero
). The only CI-significant standalone gaps are small ($\le$0.16 CLIP, the Q4\_K$-$FP8
divergence below, and $\le$0.055 Pick~
). Under the JSON prompts the model expects, \emph{none of the three standalone judges
separates the variants by a practically meaningful margin}.

\begin{table}[t]
\centering
\caption{Standalone quality over the 300-prompt benchmark under JSON prompts. All four
variants are within $\approx$0.004 HPSv2 (a point-estimate spread; no per-image CI was
computed); the CLIP and Pick CIs (Table~\ref{tab:ci}) show no practically meaningful
separation. Higher is better.}
\label{tab:quality}
\begin{tabular}{lccc}
\toprule
Variant & HPSv2$\uparrow$ & CLIP$\uparrow$ & Pick$\uparrow$ \\
\midrule
FP8 (ref)            & 0.2807 & 29.48 & 22.85 \\ 
NF4                  & 0.2789 & 29.63 & 22.80 \\ 
\textbf{INT8 (ours)} & 0.2763 & 29.48 & 22.84 \\ 
\textbf{Q4\_K (ours)}& 0.2783 & 29.64 & 22.79 \\ 
\bottomrule
\end{tabular}
\end{table}

\begin{table}[t]
\centering
\caption{Paired bootstrap 95\% CIs (10k resamples, same-seed, $n{=}300$). A CI excluding
zero indicates a difference resolved at this sample size; CIs are uncorrected for
multiple comparisons, so the few CI-significant gaps are read descriptively. The
standalone-quality CLIP CIs all include zero (or, for Q4\_K$-$FP8, are small), i.e.\ the
judges do not separate the variants.}
\label{tab:ci}
\begin{tabular}{lcc}
\toprule
Comparison & CLIP $\Delta$ (95\% CI) & Pick $\Delta$ (95\% CI) \\
\midrule
INT8 $-$ FP8 & $-0.005$ $[-0.146,+0.133]$ & $-0.007$ $[-0.040,+0.027]$ \\ 
INT8 $-$ NF4 & $-0.148$ $[-0.306,+0.008]$ & $+0.046$ $[+0.005,+0.086]$ \\ 
Q4\_K $-$ NF4 & $+0.011$ $[-0.162,+0.191]$ & $-0.003$ $[-0.046,+0.040]$ \\ 
NF4 $-$ FP8 & $+0.143$ $[-0.015,+0.299]$ & $-0.052$ $[-0.092,-0.012]$ \\ 
Q4\_K $-$ FP8 & $+0.154$ $[+0.014,+0.298]$ & $-0.055$ $[-0.093,-0.017]$ \\ 
\bottomrule
\end{tabular}
\end{table}

\paragraph{Where the variants separate: fidelity to FP8.} Standalone scores being tied,
the discriminating axis is how closely each variant reproduces the FP8 output.
Table~\ref{tab:fidelity} reports PSNR/SSIM/LPIPS against the FP8 images. \textbf{INT8 is the
most faithful} (LPIPS $0.243$, PSNR $18.53$, SSIM $0.712$
), ahead of Q4\_K ($0.277$
) and NF4 ($0.306$
); the INT8$-$Q4\_K LPIPS gap is the one fidelity comparison we tested for significance and
its CI excludes zero ($\Delta{=}{-}0.034$, CI $[-0.041,-0.027]$
), while the larger INT8$<$NF4 gap (0.243 vs.\ 0.306) rests on the point estimates. The
fidelity ordering INT8 $<$ Q4\_K $<$ NF4 (lower LPIPS = closer to FP8
) is the cleanest separation in the study and is the right way to read INT8's value: not
``higher standalone quality'' but ``the closest 8-bit reproduction of the FP8 reference'', which, being an 8-bit method tracking an 8-bit reference, is partly expected (\S\ref{sec:limitations}).

\begin{table}[t]
\centering
\caption{Reference fidelity vs.\ the FP8 images (JSON prompts, $n{=}300$). Higher
PSNR/SSIM and lower LPIPS = closer to FP8. INT8 is the most faithful; its LPIPS edge over
Q4\_K excludes zero (Table~\ref{tab:ci} method).}
\label{tab:fidelity}
\begin{tabular}{lccc}
\toprule
Variant & PSNR$\uparrow$ & SSIM$\uparrow$ & LPIPS$\downarrow$ \\
\midrule
\textbf{INT8 (ours)}  & \textbf{18.53} & \textbf{0.712} & \textbf{0.243} \\ 
Q4\_K (ours)          & 17.71 & 0.677 & 0.277 \\ 
NF4                   & 16.82 & 0.642 & 0.306 \\ 
\bottomrule
\end{tabular}
\end{table}

\begin{figure}[t]
\centering
\includegraphics[width=\linewidth]{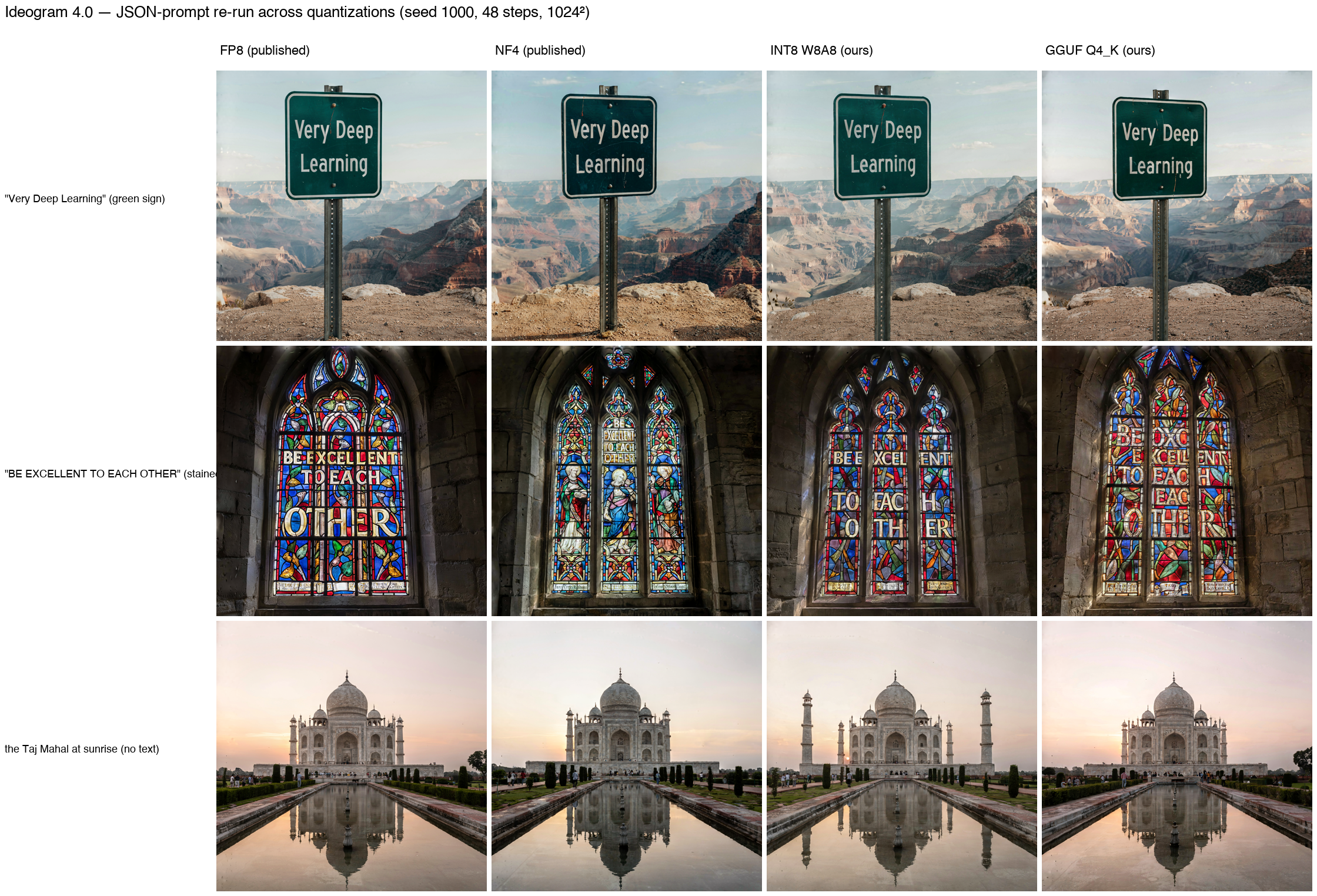}
\caption{Qualitative comparison under JSON prompts (illustrative; three captions at fixed
seed, FP8 reference vs.\ NF4 / INT8 / Q4\_K). In-image text renders legibly across all
variants, the legibility that is near-zero under raw prompts (\S\ref{sec:results}), and
the variants are visually close, consistent with the standalone-quality parity in
Table~\ref{tab:quality}. The fidelity differences (Table~\ref{tab:fidelity}) are subtle at
this scale.}
\label{fig:qualitative}
\end{figure}

\paragraph{Text rendering is real, but only under JSON prompts.} Table~\ref{tab:text}
reports OCR over the 63 text-target prompts. Exact-match is $\approx$54--57\% across
variants~
; under raw prompts the same model scored $\approx$0\%
. The gap between $\approx$0\% and $\approx$55\% is driven by the prompt protocol, not by
any variant or recipe
: text legibility is a property of how the model is prompted, not (as a raw-prompt
evaluation would conclude) a limitation of the base model. INT8 has the lowest aggregate NED
($0.099$
), but the per-variant NED differences are within noise at this sample size (INT8$-$NF4 NED
CI includes zero, $n{=}63$
); the robust statement is that all 8-bit-and-4-bit variants render text legibly under
JSON prompts (Fig.~\ref{fig:qualitative}).

\begin{table}[t]
\centering
\caption{Text rendering under JSON prompts (OCR over 63 targets, PP-OCR). Exact-match
$\approx$0\% under raw prompts~
$\rightarrow$ $\approx$55\% here; per-variant NED differences are within noise ($n{=}63$).}
\label{tab:text}
\begin{tabular}{lcc}
\toprule
Variant & OCR exact$\uparrow$ & OCR NED$\downarrow$ \\
\midrule
FP8 (ref)             & 0.540 & 0.133 \\ 
NF4                   & 0.556 & 0.116 \\ 
\textbf{INT8 (ours)}  & 0.571 & 0.099 \\ 
Q4\_K (ours)          & 0.540 & 0.123 \\ 
\bottomrule
\end{tabular}
\end{table}

\paragraph{What drives INT8 recovery: protection $>$ smoothing.} An activation-quant
ablation isolates the two levers of the recipe. Naive W8A8 (no smoothing, no protection)
collapses; SmoothQuant alone and protection alone each recover most of the gap, and
protection accounts for $\approx$78--92\% of the recovery, with the two effects
sub-additive
. The protection-size ladder shows a sharp knee, $N{=}17$ holds the ceiling while $N{=}8$
collapses back toward the unprotected baseline~
down-projections in bf16. We note this ablation was calibrated and scored on raw prompts,
on which the fragility ranking is timestep-stable
; we did not re-run the $2{\times}2$ ablation on the JSON battery (\S\ref{sec:limitations}),
so the attribution is reported for the raw-prompt calibration regime and the protected INT8
recipe (not the ablation arms) is what Tables~\ref{tab:quality}--\ref{tab:fidelity} evaluate.

\paragraph{GGUF: Q4\_K is the Pareto choice on memory.} Q4\_K (4.5\,bpw, 10.44\,GB, the
NF4 size class~
) matches NF4 on standalone quality (Q4\_K$-$NF4 CLIP and Pick CIs both include zero
) at an essentially identical on-disk size ($+0.04$\,GB), so it is the \textbf{Pareto
choice} on the quality--memory frontier. Q8\_0 is lossless by round-to-nearest
.

\paragraph{Efficiency.} Table~\ref{tab:efficiency} consolidates size and latency on the
RTX~3090 target. The two axes pull apart: INT8 matches FP8's footprint (8-bit weights),
while Q4\_K halves it; but Q4\_K is the slowest variant (its weight-only dequant runs in a
Torch kernel), and INT8 lacks a fused GEMM so it too runs eager
. No variant dominates on both memory and latency: the memory win (Q4\_K) and the
fidelity win (INT8) sit on different points of the trade-off.

\begin{table}[t]
\centering
\caption{Efficiency on the RTX~3090 target (48 steps, $1024^2$, validated
single-GPU-compute recipe). Size is on-disk DiT weights (both branches); latency is in-VRAM
s/image. INT8 is FP8-class in size and has no fused kernel; Q4\_K is half the size but the
slowest.}
\label{tab:efficiency}
\begin{tabular}{lcc}
\toprule
Variant & Size (GB)$\downarrow$ & s/img$\downarrow$ \\
\midrule
FP8 (ref)            & 18.6 & 172.9 \\ 
NF4                  & 10.4 & 164.5 \\ 
\textbf{INT8 (ours)} & 18.6 & $\sim$184 \\ 
\textbf{Q4\_K (ours)}& 10.44 & 203.3 \\ 
\bottomrule
\end{tabular}
\end{table}

\section{Discussion}
\label{sec:discussion}
The study's organizing finding is that \emph{the measurement axis decides the ranking}. On
the standalone judges that quantization papers most often report (CLIP, PickScore, and
even the stronger human-preference HPSv2), the four variants are at parity under the
JSON prompts the model expects: an 8-bit recipe, a 4-bit k-quant, and a 4-bit NF4 baseline
are indistinguishable. The prompt format is what makes this visible: evaluating the same
variants under raw natural-language prompts, CLIP reports INT8 ahead of NF4 by $\approx$2
and Q4\_K ahead by $\approx$3.6~
A study that ranked these variants on a single CLIP-family score, prompted off-distribution,
would report a quality ordering that the model's intended prompts do not support.

What \emph{does} separate the variants is reference fidelity. INT8 reproduces the
full-precision output most closely (LPIPS $0.243$, CI-significantly below Q4\_K), which is
the precise sense in which 8-bit W$+$A ``preserves the model'': not by scoring higher on a
no-reference judge, but by drifting least from the reference outputs. This reframes the
practical recommendation. If the goal is to substitute for the full-precision model with
minimal behavioral change, INT8 is the choice; if the goal is the smallest checkpoint at
equal standalone quality, Q4\_K is the choice; the two are not competitors on the same axis.

The second lesson is about text. Ideogram~4.0's signature capability, in-image text, is invisible to an evaluation that prompts it with raw strings, where OCR exact-match sits
near zero and would be mistaken for a model limitation. Under JSON prompts the same model
renders text legibly ($\approx$55\% exact-match), and the legibility is preserved across
all quantized variants. For a JSON-prompted generator, the prompt format is not an
implementation detail of the harness; it is a precondition for the text axis to exist at
all.

\section{Limitations}
\label{sec:limitations}
\paragraph{INT8's gains are quality and fidelity, not yet compute.} At 8-bit weights INT8 is
FP8-class in size ($18.6$\,GB vs.\ NF4's $10.4$~
) and, because the \texttt{ideogram4} stack ships no fused INT8 GEMM, it runs eager at
$\approx$184\,s/image~
kernels
. INT8 is best read as a quality-preserving \emph{substrate}: a fused Ampere-INT8 kernel
(CUTLASS/ViDiT-Q-style~\cite{2305.10657v4,2406.02540v3}) would convert its 8-bit compute
into a realized speed-up. Q4\_K already delivers the memory win today, and only Q4\_K (not
INT8) meets the ``$\le$ NF4 memory'' half of our success criterion
.

\paragraph{Threats to validity.}
\emph{Construct.} Standalone judges (HPSv2/CLIP/Pick) are learned-preference proxies and
are weak on fine typography; our parity claim is a claim about these judges, and a different
judge could in principle separate variants the present battery cannot. The HPSv2 spread
($\approx$0.004) carries no per-image CI, so we read it as parity, not as a ranking.
\emph{Internal.} The INT8 recipe's activation calibration and the protection-size ablation
were computed on \emph{raw} prompts; the protection set is module-level and we expect it to
be format-robust, but we did not re-run the ablation under JSON prompts. \emph{External.}
Headline statistics use a single seed (1000) over 300 prompts with paired CIs, but not
across seeds; OCR significance is limited by $n{=}63$ targets; all latency/VRAM figures come
from one RTX~3090 cluster and were not replicated on independent hardware. \emph{Reference.}
Our reference is FP8 (no public BF16 release), so ``fidelity to FP8'' is a proxy for full
precision, and Q4\_K's small standalone-score advantage over FP8 reflects model divergence
(it deviates more in pixels), not higher fidelity
.

\section{Availability}
The \textbf{quantized weights} are released on Hugging Face under a gated license consistent
with the upstream Ideogram~4.0 \emph{non-commercial, research-only} terms
:
\begin{itemize}
  \item INT8 W8A8: \url{https://huggingface.co/transformerlab/ideogram-4-int8-w8a8}
  \item GGUF Q4\_K: \url{https://huggingface.co/transformerlab/ideogram-4-gguf-q4_k}
\end{itemize}
All results use seed 1000; the evaluation recipe is available on request.

\section{Conclusion}
A SmoothQuant-plus-fragility-protection INT8 W8A8 recipe lets a 9.3B dual-branch
flow-matching DiT run at 8-bit weights and activations while holding the FP8 standalone
ceiling and reproducing the FP8 outputs more faithfully than any 4-bit baseline; a GGUF
Q4\_K encoder matches the NF4 baseline's quality at NF4's size and is the Pareto choice on
memory. The broader result is methodological: when a model is prompted the way it was
trained, the learned-preference judges we tested (CLIP, PickScore, HPSv2) stop separating
these quantized variants, while the meaningful differences move to reference fidelity and
text, so an evaluation that ignores either the prompt format or the fidelity axis will
rank these variants on differences these judges cannot actually resolve. That the variants
\emph{do} differ on fidelity and text confirms the non-separation is a property of the
judges, not evidence that the variants are identical. The clearest next steps are a fused Ampere-INT8 GEMM to turn the validated 8-bit
compute into a latency win~\cite{2305.10657v4,2406.02540v3}, re-calibrating the INT8 recipe
under JSON prompts, and step-caching for flow matching~\cite{2605.16789v1}, each measured
under the same JSON-prompt protocol.

\section*{Acknowledgements}
We thank the Ideogram team for their feedback on our results, which improved the evaluation
protocol and analysis presented in this work.

\bibliographystyle{unsrt}
\bibliography{references}

\begin{thebibliography}{10}

\bibitem{ideogram-4-2026}
Ideogram AI.
\newblock {Ideogram 4}.
\newblock \url{https://ideogram.ai/blog/ideogram-4.0/}, 2026.

\bibitem{2212.09748v2}
William Peebles and Saining Xie.
\newblock {Scalable Diffusion Models with Transformers}.
\newblock arXiv:2212.09748 [cs.CV], 2022.

\bibitem{2210.02747v2}
Yaron Lipman, Ricky T.~Q. Chen, Heli Ben-Hamu, Maximilian Nickel, and Matt Le.
\newblock {Flow Matching for Generative Modeling}.
\newblock arXiv:2210.02747 [cs.LG], 2022.

\bibitem{2211.10438v7}
Guangxuan Xiao, Ji~Lin, Mickael Seznec, Hao Wu, Julien Demouth, and Song Han.
\newblock {SmoothQuant: Accurate and Efficient Post-Training Quantization for
  Large Language Models}.
\newblock arXiv:2211.10438 [cs.CL], 2022.

\bibitem{2208.07339v2}
Tim Dettmers, Mike Lewis, Younes Belkada, and Luke Zettlemoyer.
\newblock {LLM.int8(): 8-bit Matrix Multiplication for Transformers at Scale}.
\newblock arXiv:2208.07339 [cs.LG], 2022.

\bibitem{2406.02540v3}
Tianchen Zhao, Tongcheng Fang, Haofeng Huang, Enshu Liu, Rui Wan, Widyadewi
  Soedarmadji, Shiyao Li, Zinan Lin, Guohao Dai, Shengen Yan, Huazhong Yang,
  Xuefei Ning, and Yu~Wang.
\newblock {ViDiT-Q: Efficient and Accurate Quantization of Diffusion
  Transformers for Image and Video Generation}.
\newblock arXiv:2406.02540 [cs.CV], 2024.

\bibitem{2411.05007v4}
Muyang Li, Yujun Lin, Zhekai Zhang, Tianle Cai, Xiuyu Li, Junxian Guo, Enze
  Xie, Chenlin Meng, Jun-Yan Zhu, and Song Han.
\newblock {SVDQuant: Absorbing Outliers by Low-Rank Components for 4-Bit
  Diffusion Models}.
\newblock arXiv:2411.05007 [cs.CV], 2024.

\bibitem{2503.15465v3}
Ruichen Chen, Keith~G. Mills, and Di~Niu.
\newblock {FP4DiT: Towards Effective Floating Point Quantization for Diffusion
  Transformers}.
\newblock arXiv:2503.15465 [cs.CV], 2025.

\bibitem{2305.10657v4}
Yefei He, Luping Liu, Jing Liu, Weijia Wu, Hong Zhou, and Bohan Zhuang.
\newblock {PTQD: Accurate Post-Training Quantization for Diffusion Models}.
\newblock arXiv:2305.10657 [cs.CV], 2023.

\bibitem{2501.00124v1}
Jiaojiao Ye, Zhen Wang, and Linnan Jiang.
\newblock {PQD: Post-training Quantization for Efficient Diffusion Models}.
\newblock arXiv:2501.00124 [cs.CV], 2024.

\bibitem{2412.06661v2}
Shuaiting Li, Juncan Deng, Zeyu Wang, Kedong Xu, Rongtao Deng, Hong Gu, Haibin
  Shen, and Kejie Huang.
\newblock {Efficiency Meets Fidelity: A Novel Quantization Framework for Stable
  Diffusion}.
\newblock arXiv:2412.06661 [cs.CV], 2024.

\bibitem{2605.16732v1}
Sayeh Sharify, Mahsa Salmani, and Hesham Mostafa.
\newblock {DiRotQ: Rotation-Aware Quantization for 4-bit Diffusion
  Transformers}.
\newblock arXiv:2605.16732 [cs.CV], 2026.

\bibitem{2210.17323v2}
Elias Frantar, Saleh Ashkboos, Torsten Hoefler, and Dan Alistarh.
\newblock {GPTQ: Accurate Post-Training Quantization for Generative Pre-trained
  Transformers}.
\newblock arXiv:2210.17323 [cs.LG], 2022.

\bibitem{2306.00978v6}
Ji~Lin, Jiaming Tang, Haotian Tang, Shang Yang, Wei-Ming Chen, Wei-Chen Wang,
  Guangxuan Xiao, Xingyu Dang, Chuang Gan, and Song Han.
\newblock {AWQ: Activation-aware Weight Quantization for LLM Compression and
  Acceleration}.
\newblock arXiv:2306.00978 [cs.CL], 2023.

\bibitem{2305.14314v1}
Tim Dettmers, Artidoro Pagnoni, Ari Holtzman, and Luke Zettlemoyer.
\newblock {QLoRA: Efficient Finetuning of Quantized LLMs}.
\newblock arXiv:2305.14314 [cs.LG], 2023.

\bibitem{2409.00492v1}
Vage Egiazarian, Denis Kuznedelev, Anton Voronov, Ruslan Svirschevski, Michael
  Goin, Daniil Pavlov, Dan Alistarh, and Dmitry Baranchuk.
\newblock {Accurate Compression of Text-to-Image Diffusion Models via Vector
  Quantization}.
\newblock arXiv:2409.00492 [cs.CV], 2024.

\bibitem{2605.27003v1}
Junhao Wu, Dezhong Yao, and Hai Jin.
\newblock {Timestep-Aware SVDQuant-GPTQ for W4A4 Quantization of Wan2.2-I2V}.
\newblock arXiv:2605.27003 [cs.CV], 2026.

\bibitem{2606.00957v1}
Yiming Zhao.
\newblock {Boundary-Protection W8A8 HiFloat8 Quantization for Large-Scale
  Text-to-Video Diffusion Transformers}.
\newblock arXiv:2606.00957 [cs.CV], 2026.

\bibitem{2411.19108v2}
Feng Liu, Shiwei Zhang, Xiaofeng Wang, Yujie Wei, Haonan Qiu, Yuzhong Zhao,
  Yingya Zhang, Qixiang Ye, and Fang Wan.
\newblock {Timestep Embedding Tells: It's Time to Cache for Video Diffusion
  Model}.
\newblock arXiv:2411.19108 [cs.CV], 2024.

\bibitem{2605.16789v1}
Xiao Liu, Kai Liu, Naiyang Guan, Hongliang Lu, Zhixin Wang, Zhikai Chen,
  Renjing Pei, and Yulun Zhang.
\newblock {Accelerating Rectified Flow Models via Trajectory-Aware Caching}.
\newblock arXiv:2605.16789 [cs.CV], 2026.

\bibitem{2312.00858v2}
Xinyin Ma, Gongfan Fang, and Xinchao Wang.
\newblock {DeepCache: Accelerating Diffusion Models for Free}.
\newblock arXiv:2312.00858 [cs.CV], 2023.

\bibitem{2602.01273v4}
Xun Zhang, Kaicheng Yang, Hongliang Lu, Haotong Qin, Yong Guo, and Yulun Zhang.
\newblock {Q-DiT4SR: Exploration of Detail-Preserving Diffusion Transformer
  Quantization for Real-World Image Super-Resolution}.
\newblock arXiv:2602.01273 [cs.CV], 2026.

\bibitem{2412.17225v1}
Lichen Ma, Tiezhu Yue, Pei Fu, Yujie Zhong, Kai Zhou, Xiaoming Wei, and Jie Hu.
\newblock {CharGen: High Accurate Character-Level Visual Text Generation Model
  with MultiModal Encoder}.
\newblock arXiv:2412.17225 [cs.CV], 2024.

\bibitem{2412.00136v3}
Wenda Shi, Yiren Song, Dengming Zhang, Jiaming Liu, and Xingxing Zou.
\newblock {FonTS: Text Rendering with Typography and Style Controls}.
\newblock arXiv:2412.00136 [cs.CV], 2024.

\bibitem{2509.01624v1}
Natalia Frumkin and Diana Marculescu.
\newblock {Q-Sched: Pushing the Boundaries of Few-Step Diffusion Models with
  Quantization-Aware Scheduling}.
\newblock arXiv:2509.01624 [cs.CV], 2025.

\end{thebibliography}

\end{document}